# Conservative Exploration for Policy Optimization via Off-Policy Policy Evaluation

**Paul Daoudi**[*][†]   **Mathias Formoso**[*][†]   **Othman Gaizi**[*][†]   **Achraf Azize**[*][†]   **Evrard Garcelon**[‡]
{paul.daoudi, mathias.formoso, othman.gaizi, achraf.azize}@ens-paris-saclay.fr
evrard.garcelon@ensae.fr

## Abstract

A precondition for the deployment of a Reinforcement Learning agent to a real-world system is to provide guarantees on the learning process. While a learning algorithm will eventually converge to a good policy, there are no guarantees on the performance of the exploratory policies. We study the problem of conservative exploration, where the learner must at least be able to guarantee its performance is at least as good as a baseline policy. We propose the first conservative provably efficient model-free algorithm for policy optimization in continuous finite-horizon problems. We leverage importance sampling techniques to counterfactually evaluate the conservative condition from the data self-generated by the algorithm. We derive a regret bound and show that (w.h.p.) the conservative constraint is never violated during learning. Finally, we leverage these insights to build a general schema for conservative exploration in DeepRL via off-policy policy evaluation techniques. We show empirically the effectiveness of our methods.

## 1 Introduction

Exploration is one of the main challenges in Reinforcement Learning (RL) and also one of the reasons why RL is rarely deployed in real systems. Exploration means that the algorithm is willing to sacrifice rewards for policies improving the knowledge of the system. In general, there is no guarantee on the performance of such exploratory policies. This is critical is applications such as digital marketing, healthcare, and robotics where hard constraints must be satisfied. The lack of guarantees on the learning process is one of the major obstacle to deploy RL to real-world systems. For many domains, it is common to have a reliable but potientially suboptimal *baseline policy*. Devising exploration strategies able to leverage this prior knowledge and perform at least as well as the baseline policy is an important challenge of RL that needs to be solved for any real-world use.

The bandit literature has studied this problem under the name of *conservative exploration*. Examples include multi-armed bandits [1, 2, 3], contextual linear bandits [3, 4, 5], and combinatorial bandits [3, 6, 7, 8]. Conservative exploration has been recently investigated in *tabular RL* in [9] where at each episode, the cumulative expected performance of the policy deployed in the environment should never be worse (with high probability) than the cumulative performance of a known baseline policy, $\pi_b$. Their algorithm, CUCBVI [9], constructs confidence intervals based on a model estimation procedure. This allows the algorithm to only deploy policies satisfying the conservative exploration constraint. However, using a model-based approach limits the algorithm capacity to handle large scale or continuous problems. Hence, it is natural to ask if model-free algorithms, that avoiding to estimate the transition model are more amenable for larger or continuous problems, can satisfy the conservative constraint. In this paper, we answer the following question:

*Is it possible to design a model-free algorithm for conservative exploration?*

---
[*]Equal contribution
[†]ENS Paris-Saclay
[‡]ENSAE Paris

This paper aims to provide a positive answer to this question. First, we provide a model-free algorithm for conservative exploration[4] with provable guarantees based on multiple importance sampling [e.g. 10, 11]. Notably this is also the first algorithm with guarantees in continuous MDPs and policy parametrization. Second, we leverage the insights from the theoretical algorithm to provide a generic framework for conservative exploration in Deep RL (DRL) based on OPE. We conclude the paper with a set of empirical validations.

**Related Work.** Safety w.r.t. a baseline has been largely studied in batch settings [12, 13, 14, 15, 16, 17, 18, 19]. Given a set of samples collected through a baseline policy, the objective of "safe" batch RL is to learn a policy that is guaranteed to improve the baseline, without interacting with the system. This implicitly requires that the samples are sufficiently rich or, in other words, that the baseline policy provides a sufficient level of exploration. This framework can be extended to semi-batch learning [e.g. 20] by alternating the phase of offline learning with the execution of the improved policy. In general, semi-batch approaches do not explicitly take into account the exploration problem, not to mention convergence guarantees. Notable exceptions are the work on conservative policy iteration in tabular MDPs [21, 22].

A related but different problem is the one of safety, where the algorithm or optimal policy is forced to satisfy a set of constraints, potentially not aligned with the reward signal [e.g. 23, 24, 25]. In the bandit literature, safety with linear constraints have been studied in [26, 27]. In RL, this problem has been modeled through constrained MDPs [28]. While conservative exploration poses a constraint on the learning algorithm, CMDPs poses a constraint on the optimal policy by requiring it to satisfy a set of constraints on signals potentially unrelated to the reward. This means the optimal policy of a CMDP may be different than the one in the unconstrained MDP. Online learning algorithms for CMDPs [e.g., 29, 30, 31, 32] focus on providing guarantees on the number of constraint violations. More practical algorithms have also been proposed in the literature [see e.g., 33, 34, 35]. In this line of research, perhaps the work most closely related to our own is [36], where the exploration is modified to prevent the agent from entering disastrous states. It is modeled through a CMDP, and solved with a combination of Conservative-Q-Learning [37] and Constrained-Policy-Optimization [20, 33]. Our work aims to further tighten those constraints with the knowledge of a baseline policy.

## 2 Notation

A Markov Decision Process $M = (\mathcal{S}, \mathcal{A}, p, r)$ is defined by a state space $\mathcal{S}$, an action space $\mathcal{A}$, a transition model $p : \mathcal{S} \times \mathcal{A} \to \Delta(\mathcal{S})$, a reward function $r : \mathcal{S} \times \mathcal{A} \to [0, r_{\max}]$ and initial state $s_1$.[5] Note that this work encompasses the case where $\mathcal{S}$ and $\mathcal{A}$ are continuous. We consider a finite-horizon problem with horizon $H \in \mathbb{N}_{>0}$.[6]

A Markov randomized policy $\pi = (d_1, \ldots, d_H) \in \Pi$ is a sequence of $H$ decision rules $d : \mathcal{S} \to \Delta(\mathcal{A})$ defining a distribution over actions. The value of a policy is measured through the value functions $Q_h^\pi(s, a) = \mathbb{E}_\pi \left[ \sum_{i=h}^H r(s_i, a_i) | s_h = s, a_h = a \right]$ and $V_h^\pi = \mathbb{E}_{a \sim \pi(\cdot|s)}[Q_h^\pi(s, a)]$, where the expectation is defined w.r.t. the model and policy. There exists an optimal policy $\pi^\star$ such that $V_h^\star = V_h^{\pi^\star}$ satisfies the optimality equations $V_h^\star(s) = \max_{a \in \mathcal{A}}\{r(s, a) + \mathbb{E}_{s'}[V_{h+1}^\star(s')]\}$ for any $s \in \mathcal{S}$ and $h \in [H]$.

We can rewrite this optimization problem in a way that is more amenable for *model-free policy optimization*. For a policy $\pi \in \Pi$, the policy performance is defined as

$$J(\pi) = V_1^\pi(s_1) = \mathbb{E}_\pi \left[ \sum_{h=1}^H r(s_h, a_h) \right] = \mathbb{E}_{\tau \sim q(\cdot|\pi)} \left[ f(\tau) \right]. \tag{1}$$

---
[4]We use the term model-free to denote an algorithm that does not estimate transitions and rewards and is amenable to be used with function approximation. However, the proposed algorithm needs to store the samples to perform off-policy policy evaluation.

[5]In the case of stochastic initial state we can extend the MDP with a new stage with single state and action such that the transition probability matches the initial state distribution.

[6]The results can be extended to infinite-horizon discounted settings with reset by paying a constant error due to the truncation of a trajectory to $H$ steps. In the extension to deepRL, we indeed consider discounted infinite-horizon problems.

where $q(\cdot|\pi)$ is the density function of a trajectory $\tau = (s_h, a_h, r_h)_{h \in [H]}$ under policy $\pi$, and $f(\tau) = \sum_{h=1}^{H} r_h$ is the cumulative reward of a trajectory. Then, the optimal policy is such that $\pi^\star \in \arg\max_{\pi \in \Pi} J(\pi)$. The set $\Pi$ characterizes the class of policies we are interested in. For example, action-based policy methods [e.g. 38] consider a parametric policy class $\Pi_\theta = \{\pi_\theta : \theta \in \Theta \subseteq \mathbb{R}^d\}$, while parameter-based methods [e.g. 39] a so-called hyperpolicy class $\Pi_{\xi,\theta} = \{\pi_\theta : \theta \sim \nu_\xi, \xi \in \Xi\}$. Refer to [40, 41] for more details on policy optimization.

## 3 Conservative Exploration in Policy Optimization

We consider the standard online learning protocol in finite-horizon problems. The agent interacts with an unknown environment in a sequence of episodes. At each episode $k$, the learner executes a policy $\pi_k$, records the samples $\tau_k = (s_{kh}, a_{kh}, r_{kh})_{h \in [H]}$ of the trajectory, and updates the policy. In conservative exploration, a learning agent is expected to perform as well as the optimal policy over time (in a regret minimization sense), under the constraint that at no point in time its performance is significantly worse than the one of a baseline policy $\pi_b$. For simplicity, we assume that both $\pi_b$ and $V^{\pi_b}$ are known. In general, this assumption is not restrictive since the baseline performance can be estimated from historical data. We will however relax this assumption later in this section.

The conservative condition for MDPs was first introduced in [9]. Given a level $\alpha \in (0, 1)$, the conservative condition is:

$$\forall k > 0, \quad \sum_{l=1}^{k} J(\pi_l) \geq (1-\alpha) \sum_{l=1}^{k} J(\pi_b), \quad \text{w.h.p.} \qquad (2)$$

If (2) is verified, the (exploratory) policy $\pi_k$ is run at episode $k$, otherwise the baseline policy $\pi_b$ is played. Note that $J(\pi_l)$ cannot be computed exactly since the MDP is unknown. It is thus necessary to estimate it using the trajectories $D_k = \bigcup_{l=1}^{k-1} \tau_l$ collected before episode $k$. In order to be conservative and guarantee that the original condition holds, it is necessary to build a pessimistic estimate of $\sum_{l=1}^{k} J(\pi_l)$. A sufficient condition is to construct an estimate $\underline{J}(\pi_l)$ such that $\underline{J}(\pi_l) \leq J(\pi_l)$, for any $l \leq k$. Since the trajectories have been collected by running multiple policies, this problem is an instance of (pessimistic) *offline policy evaluation*.

The authors of [9] solved this counterfactual estimation problem using a model-based approach. At the beginning of each episode $k$, their algorithm –called CUCBVI– builds a set of plausible MDPs defined a $\mathcal{M}_k = \{M = (\mathcal{S}, \mathcal{A}, \overline{r}, \overline{p}) : \overline{r}(s,a) \in B_r^k(s,a), \overline{p}(s'|s,a) \in B_p^k(s,a,s'), \sum_{s'} \overline{p}(s'|s,a) = 1\}$ where $B_r^k$ and $B_p^k$ are high-probability confidence intervals on the rewards and transition probabilities of the true MDP $M^\star$, which guarantee $M^\star \in \mathcal{M}_k$ w.h.p. The set $\mathcal{M}_k$ allows to build optimistic and pessimistic estimates of the value of each policy $\pi$. The optimistic estimate is used to select the policy to run while the pessimistic to evaluate the conservative condition. Formally, CUCBVI selects the exploratory (optimistic) policy $\pi_k$ by solving the problem $\max_{M \in \mathcal{M}_k} \max_\pi \{V_1^\pi(M)\}$. To decide whether to play $\pi_k$ or not, CUCBVI builds a pessimistic estimate of all the policies $\{\pi_l\}_{l \in [k]}$ by computing $\underline{V}_1^{\pi_l} = \min_{M \in \mathcal{M}_k} \{V_1^{\pi_l}(M)\}$. If $\sum_{l=1}^{k} \underline{J}(\pi_l) = \sum_{l=1}^{k} \underline{V}_1^{\pi_l}(s_1) \geq (1-\alpha) \sum_{l=1}^{k} J(\pi_b)$, the exploratory policy $\pi_k$ is safe to play. This guarantees that w.h.p. the overall performance of the algorithm is no worse than the one of the baseline policy (up to a factor $1 - \alpha$) over $k$ episodes. The main limitation of this approach is its model-based nature. Building a provably good model of the MDP is possible only in restricted cases, for example in tabular MDPs or linear mixture MDPs [e.g. 42].[7] An idea to overcome this limitation may be to build conservative exploration algorithm on top of model-free algorithms based on Q-learning [e.g. 43]. While these algorithms builds an optimistic and pessimistic estimate of the optimal Q-function $Q^\star$, they do not allow to obtain a counterfactual estimate of any arbitrary policy, thus hindering the evaluation of the conservative condition.

To overcome these limitations and design a more practical algorithm, we take a rollout-based view (i.e., trajectory-based view) of the problem. The main idea is to leverage the correlation among policies induced by the stochasticity of the MDP and policies themselves. Under certain conditions (see Sec. 3.1), we can use trajectories collected through a policy to perform a counterfactual estimate of the performance of another policy via importance sampling. While any offline policy evaluation approach can be used [see e.g., 44, 45], we leverage the idea of multiple importance sampling (MIS) in [11] to build a confidence interval around the estimated value.

---

[7] In the literature there is not a conservative algorithm for linear mixture MDPs but we think it is possible to extend the idea in [9] to this setting.

At episode $k$, we have access to the set of trajectories $D_k = \bigcup_{l=1}^{k-1} \tau_l$ collected through the policies $\{\pi_l\}_{l<k}$. We denote by $\{\mu_1, \ldots, \mu_M\}$ the set of unique policies played by the algorithm each one associated to $N_m$ trajectories $\{\tau_{mj}\}_{j\in[N_m]}$ ($\sum_m N_m = k-1$). The Robust Balance Heuristic (RBH) estimator of the performance of policy $\pi$ is

$$J_k^{\text{RBH}}(\pi) = \frac{1}{k-1} \sum_{m=1}^{M} \sum_{j=1}^{N_m} \min \left\{ C, (k-1) \underbrace{\frac{q(\tau_{mj}|\pi)}{\sum_{i=1}^{M} N_i q(\tau_{mj}|\mu_i)}}_{:=\omega_{D_k}^{\text{BH}}(\tau_{mj})} \right\} f(\tau_{mj}). \tag{3}$$

The term $C$ plays the role of controlling the trade-off between bias and variance of the estimate. If $C = +\infty$ or it is large enough, the robust estimates reduces to the standard BH that is an unbiased estimate of $J(\pi)$. In general, the RBH estimate is biased but, by reducing the range of the weights, leads to an improved variance. Using an adaptive truncation, it is possible to prove that, for some $\varepsilon \in (0, 1]$ [11, Thm. 1]

$$|J_k^{\text{RBH}}(\pi) - J(\pi)| \leq \beta_k(\pi) = O\left((k-1)^{-\frac{\varepsilon}{1+\varepsilon}}\right). \tag{4}$$

We refer the reader to App. B for a complete discussion about this concentration. Similarly to model-based approaches, we can leverage this concentration bound to build both an optimistic and a pessimistic bound to the performance of any policy $\pi \in \Pi$.[8] The main advantage is that this approach is model-free and can be used in the framework of policy optimization (i.e., with a parametric representation) and in continuous MDPs.

Similarly to OPTIMIST [11], at each episode $k$, our algorithm selects the optimistic (exploratory) policy $\pi_k$ by solving the following optimization problem

$$\max_{\pi \in \Pi} \{J_k^{\text{RBH}}(\pi) + \beta_k(\pi)\}. \tag{5}$$

where $\beta_k(\pi)$ is defined as in (4). However, before executing the policy, our algorithm verifies that $\pi_k$ is safe to play by checking the following conservative condition

$$\sum_{l=1}^{k} J_k^{\text{RBH}}(\pi_l) - \beta_k(\pi_l) \geq (1-\alpha)kJ(\pi_b). \tag{6}$$

If policy $\pi_k$ satisfies Eq. (6) then

$$\sum_{l=1}^{k} J(\pi_l) \geq (1-\alpha)kJ(\pi_b) \qquad \text{w.h.p.} \tag{7}$$

and our algorithm runs the optimistic policy, otherwise it selects the baseline policy. The resulting algorithm, called COPTIMIST, is reported in Alg. 1.

COPTIMIST performs direct exploration in the policy space by leveraging the principle of optimism-in-the-face-of-uncertainty. Due to its structure it can be interpreted as a bandit algorithm over the policy space. The fundamental difference is that COPTIMIST is able to exploit the correlation among policies instead of considering each policy independently. This allows to share samples in the estimation of the policy performance and avoid the necessity of building a model of the MDP that would be impractical in continuous domains. In fact, as shown in (8)-(9), it is not necessary to know the MDP to compute the counterfactual value of a policy through the RBH estimate. COPTIMIST *is thus the first model-free algorithm for conservative exploration.*

**Policy Space.** This flexible structure and the model-free nature allow COPTIMIST to work in different settings (see Sec. 3.1 for requirements on the policy space). In tabular settings, the policy space $\Pi$ is a set of distribution over the finite state-action space. In continuous MDPs, we can leverage a parametric policy representation defined either by a finite or compact set $\Theta$ of policy parameters. In the latter case, we have that $\Pi = \Pi_\theta$ and

$$\omega_{D_k}^{\text{BH}}(\tau, \pi_\theta) = \frac{\prod_{t=1}^{H} \pi_\theta(s_t, a_t)}{\sum_{i=1}^{M} \prod_{t=1}^{H} \pi_{\theta_i}(s_t, a_t)} \tag{8}$$

---

[8]It is known that restrictions on the behavioral policy are necessary to guarantee a meaningful off-policy evaluation [e.g., 46, 47, for soft behavioral policy or absolute continuity]. We will discuss about this in Sec. 3.1

**Algorithm 1** Conservative OPTIMIST

**Input:** baseline policy $\pi_b$
**for** $k = 1, \ldots$ **do**
    Select policy $\pi_k = \mathrm{argmax}_{\pi \in \Pi}\{J_k^{\mathrm{RBH}}(\pi) + \beta_k(\pi)\}$
    **if** (6) not verified **then**
        Set $\pi_k = \pi_b$
    **end if**
    Run policy $\pi_k$ and observe trajectory $\tau_k = (s_{kh}, a_{kh}, r_{kh}, s_{k,h+1})_{h \in [H]}$
    Set $D_{k+1} = D_k \cup \{\tau_k\}$
**end for**

This algorithmic structure encompasses also the *parameter-based* policy optimization paradigm which introduces a distribution $\nu$ over policy parameters. In this case, the optimization becomes to find the optimal hyperpolicy $\nu^\star$: $\max_{\xi \in \Xi} J(\xi) = \max_{\xi \in \Xi} \mathbb{E}_{\theta \sim \nu_\xi}[J(\theta)]$. For the parameter-based policy optimization problem, we can instantiate COPTIMIST with $\Pi = \Pi_{\xi,\theta}$ and $q(\tau | \pi = (\xi, \theta)) = \nu_\xi(\theta) q(\tau | \theta)$, where $\theta \sim \nu_\xi$. As a consequence,

$$\omega_k^{\mathrm{BH}}(\tau, \pi) = \frac{\nu_\xi(\theta)}{\sum_{i=1}^M N_i \nu_{\xi_i}(\theta)} \quad (9)$$

**Unknown baseline value.** If $J(\pi_b)$ is unknown, we can use the same counterfactual approach to evaluate the performance of the baseline policy. To guarantee that condition (7) holds w.h.p., we need to change the condition checked by the algorithm to account for the uncertainty in the baseline estimate. At episode $k$, we compute an upper-bound $\overline{J}_k(\pi_b) = J_k^{\mathrm{RBH}}(\pi_b) + \beta_k(\pi_b)$ using (3) and (4), and we evaluate the condition $\sum_{l=1}^k \underline{J}_k(\pi_j) \geq (1-\alpha)k\overline{J}_k(\pi_b)$.

### 3.1 Theoretical Guarantees

As commonly done in online learning, we evaluate the performance of an algorithm through the regret: $R(K) = \sum_{k=1}^K J^\star - J(\pi_k)$ where $J^\star = \max_{\pi \in \Pi} J(\pi)$. We consider both the case of a discrete set of policies (i.e., $|\Pi| < +\infty$) and a compact set of policy parametrizations (Asm. 3). We define by $\Delta_b = J^\star - J(\pi_b)$ the suboptimality gap of the baseline policy $\pi_b$. We introduce the following assumptions necessary for the analysis.

**Assumption 1.** *The baseline policy is such that $\Delta_b > 0$.*

**Assumption 2.** *The following bound on the $(1 + \epsilon)$-Rényi divergence holds for any $k \in [K]$:* $v_\epsilon := \sup_{\pi_1, \ldots, \pi_k \in \Pi} d_{1+\epsilon}\left(q(\cdot | \pi_k) \| \frac{1}{k-1} \sum_{j=1}^{k-1} q(\cdot | \pi_j)\right) < \infty.$

**Assumption 3.** *Let $\Pi$ be a compact set contained in a box $[-D, D]^d$, with $D > 0$. The performance $J(\pi)$ is L-Lipschitz continuous, i.e., $\forall \pi, \pi' \in \Pi$: $|J(\pi) - J(\pi')| \leq L\|\pi - \pi'\|_1$*

Asm. 1 is simply to rule out the trivial case that the baseline is already optimal. If $\Delta_b = 0$ condition (6) is never verified w.h.p. for small enough $\alpha$ values. It is common in the counterfactual evaluation literature to have assumptions on the behavior policy and evaluation policy to guarantee a meaningful estimate [e.g. 46, 47]. In our setting (online learning), we need to require a similar assumption on the entire policy space since we are not aware a priori of the policies selected by the algorithm. This intuition is formally stated in Asm. 2. This assumption can be enforced through the design of the policy space, see [11, App. B].

Denote by $\Lambda_K$ the set of episodes where an optimistic policy $\pi_k$ is played. The set $\Lambda_K^c$ is its complement. We can now state the regret bound of COPTIMIST.

**Theorem 3.1.** *Under Asm. 1, 2 and 3, for any $K$ and conservative level $\alpha$, there exists a numerical constant $\beta > 0$ such that the regret of COPTIMIST is bounded as*

$$R(K) \leq \beta R_{\mathit{OPTIMIST}}(K | \Lambda_k) + \beta \frac{\Delta_b}{\alpha J(\pi_b)} \Big( \underbrace{\varepsilon(1 + \varepsilon) \frac{(\nu_\varepsilon L_K(\delta))^{1+1/\varepsilon}}{(\Delta_b + \alpha \mu_b)^{1/\varepsilon}}}_{:= \text{\textcircled{B}}} + H + \underbrace{\frac{LD\pi^2}{6}}_{\text{\textcircled{A}}} \Big) \quad (10)$$

*and $\sum_{l=1}^k J(\pi_l) \geq (1-\alpha)k J(\pi_b)$ is verified at every $k \in [K]$ with probability $1 - \delta$. Note that $L_K(\delta)$ is a logarithmic term in $K$. Under Asm. 3 we have that $L_K(\delta) = \log\left(\frac{\pi^2 K^2}{6\delta}\right) + d \log\left(1 + dK^2\right)$.*

The term $R_{\mathrm{OPTIMIST}}(K | \Lambda_K)$ denotes the regret of OPTIMIST over the episodes where an optimistic policy was played. The second term represents the regret incurred in the conservative episode $\Lambda_k^c$ where $\pi_b$ was played. We can notice that this term has only a logarithmic dependence on the number of episodes $K$. In [11], the authors provided a regret bound of $\widetilde{O}(K^{\frac{1}{1+\epsilon}})$ and $\widetilde{O}(d^{\frac{\epsilon}{1+\epsilon}}K^{\frac{1}{1+\epsilon}})$ for discrete and compact policy space. Thus, Thm. 3.1 shows that COPTIMIST is able to satisfy the conservative condition without compromising the guaranteed on the learning performance.

The same results hold for a discrete policy space by setting Ⓐ $= 0$ and $L_K(\delta) = \log(\pi^2 K^2/6\delta)$. In the case of compact policy space (Asm. 3), solving the optimization problem Eq. (5) may be difficult. We can show a similar result to Thm. 3.1 for a variant of COPTIMIST that uses progressively finer discretization of the policy space. We refer the reader to App. D.3 for further details.

**Proof sketch.** We provide an intuition of the proof of Thm. 3.1.s We can decompose the episodes into the episodes associated to the optimistic algorithm $\Lambda_k$ and the conservative episodes $\Lambda_k^c$. As a consequence, the regret is decomposed as $R(T) \lesssim R(\text{OPTIMIST}|\Lambda_k) + |\Lambda_k^c|\Delta_b$. Under Asm. 3, the first term can be bounded by $\widetilde{O}(d^{\frac{\epsilon}{1+\epsilon}}K^{\frac{1}{1+\epsilon}})$ using the standard analysis of OPTIMIST. What is left is to show that the number of times the baseline policy is selected is sub-linear. In App. E, we show that there exists an episode $K_\varepsilon \approx \frac{Ⓑ}{\alpha J(\pi_b)}$ such that for any $l > K_\varepsilon$, COPTIMIST always plays the optimistic policy. This leads to the result in Thm. 3.1.

Combining the results, we obtain a regret of $\widetilde{O}(K^{\frac{1}{1+\epsilon}})$ for discrete policy set and $\widetilde{O}(d^{\frac{\epsilon}{1+\epsilon}}K^{\frac{1}{1+\epsilon}})$ for compact policy space that match the regret of OPTIMIST [11, 48]. This shows that COPTIMIST is able to satisfy the conservative condition without compromising the guaranteed on the learning performance. In addition, when choosing $\varepsilon = 1$, we obtain a regret of order $\widetilde{\mathcal{O}}(\sqrt{K} + \frac{\Delta_b}{\alpha J(\pi_b)(\alpha J(\pi_b)+\Delta_b)})$, a regret similar to the previous results in tabular MDPs [9].

## 4 Extension To Deep RL

In Section 3, we proposed a model-free conservative algorithm with a sub-linear regret able to respect the conservative condition with high-probability. While having theoretical guarantees on environments with a continuous state-action space, COPTIMIST does not scale to high-dimensional problems. First, the optimization step (5) can be cumbersome depending on the complexity of the policy space. Second, getting an upper bound of the $\alpha$-Rényi divergence for computing the bonus $\beta_k(\pi)$ in (4) may be intractable in practice.[9] Finally, COPTIMIST requires a significant correlation between policy trajectories to effectively leverage OPE via importance sampling and avoid an exhaustive exploration of the policy space (see Asm. 2). Experimental evidences are available in App. E.1. Despite these limitations, COPTIMIST provides an appealing structure that can be leveraged to design more practical algorithms.

**Algorithm 2** Conservative Agent through OPE
---
**Input:** baseline policy $\pi_b$, learning agent $\mathfrak{A}_l$, off-line evaluation agent $\mathfrak{A}_e$
Run policy $\pi_b$ and collect samples in $D_1$
**for** $k = 1, \ldots$ **do**
    $\mathfrak{A}_l$ selects policy $\pi_k$ using $D_k$
    Use $\mathfrak{A}_e$ to build a pessimistic estimate of (7) using $D_k$ and $\{\pi_j\}_{j \in [k]}$
    **if** condition not verified **then**
        Set $\pi_k = \pi_b$
    **end if**
    Run policy $\pi_k$ and observe trajectory $\tau_k = (s_{kh}, a_{kh}, r_{kh})_{h \in [H]}$
    Set $D_{k+1} = D_k \cup \{\tau_k\}$
**end for**

The building blocks of a conservative algorithm are: **B1)** a (non-conservative) learning agent $\mathfrak{A}_l$; **B2)** an OPE algorithm $\mathfrak{A}e$; and **B3)** a way of measuring uncertainty of the estimate values. The generic schema of a conservative algorithm is reported in Alg. 2. The *learning agent* can be any algorithm for control (i.e., aiming to learn the optimal policy). In principle, its should be able to learn from off-policy data to efficiently leverage the samples collected through the baseline policy. While this is not a strict requirement, it may be important to improve the sample complexity of the algorithm. The *OPE algorithm* is devoted to the counterfactual estimate of the policy performance, necessary to evaluate the conservative condition (7). This component can be decoupled from the learning algorithm. Several OPE methods have been studied in the literature, we refer the reader to App. A for a review.

These methods come with different guarantees and requirements that make them more or less suited for our setting. In fact, similarly to what is done in (6), we need to build a pessimistic estimate of the policy performance to guarantee that the conservative condition is not violated. This requires to be able to evaluate the uncertainty of the OPE estimate. While there are notable exceptions [49, 50, 51], the majority of the approaches do not provide provable guarantees on their estimates.

---
[9]As shown in [11], an upper-bound of the $\alpha$-Rényi divergence can be computed between any mixture of Gaussian distributions

The first idea is to use OPE methods based on importance sampling to build an estimate of the policy performance. However, the curse of horizon[10] and the fact that it is often necessary to retain the behavioral policies, make IPS methods not well suited for large DRL approaches.

Direct methods [e.g. 55] either estimates the value function or the state visitation problem, often through a primal-dual formulation [e.g. 56, 57, 58]. Under certain condition on the data generating process and function approximation, it is also possible to build confidence intervals around the OPE estimate [45]. Direct methods approaches have become more popular in the recent years due to their scalability to function approximation. We propose to leverage OPE direct approaches to overcome IPS limitations and design a practical conservative DRL algorithm.

We now provide an generic framework for conservative exploration in DRL. **B1)** We consider any learning algorithm (e.g., DQN, SAC, REINFORCE, etc.). **B2)** We consider the Fitted Q-Evaluation [59] algorithm. FQE is a model-free OPE direct method that aims to build a parametric approximation of the value function $Q^\pi$ from a set of trajectories collected via behavioral policies. For finite-horizon problems, given a target policy $\pi$, FQE iteratively applies (with $\widehat{Q}^\pi_{H+1} = 0$):

$$\widehat{Q}^\pi_h = \min_{g \in \mathcal{G}} \frac{1}{N} \sum_{i=1}^{n} \left( g(s^i_h, a^i_h) - r^i_h - \mathbb{E}_{a' \sim \pi}[\widehat{Q}_{h+1}(s^i_{h+1}, a')] \right)^2 \quad (11)$$

where $\mathcal{G}$ is an arbitrary function space.

In [60], the authors provide a data dependent confidence bounds for FQE with linear function approximation. FQE is a very appealing algorithm that is at the core of many practical algorithms, especially in continuous control [61, 62, 48, 63]. **B3)** To estimate the uncertainty, we mainly investigated two similar approaches. The first one is known as statistical bootstrapping [64, 65] and consists in creating $b$ bootstrapped estimates $\hat{v}^k_\pi$ with FQE using $b$ boostrapped datasets sampled from the entire dataset. We can then construct the $1 - \delta$ asymptotic confidence interval of the policy value by: CI$(\delta) = [\hat{v}_\pi - q^\pi_{1-\delta/2}, \hat{v}_\pi - q^\pi_{\delta/2}]$ where $\hat{v}_\pi$ is the policy value FQE estimate using the entire dataset, and $q^\pi_{1-\delta/2}, q^\pi_{\delta/2}$ are $1 - \delta/2$ and $\delta/2$ quantiles of $\{\hat{v}^k_\pi - \hat{v}_\pi\}^b_{k=1}$. Hence, the conservative condition we aim to respect at each episode start is:

$$\sum_{t=1}^{T} (\hat{v}_{\pi_t} - q^{\pi_t}_{1-\delta/2}) \geq (1 - \alpha) \sum_{t} v_{\pi_b} \quad (12)$$

Another approach, which however comes without theoretical confidence intervals, is to use Ensemble Learning, which has been used in RL [66, 67] for improving both exploration [68, 69] and learning [70, 71]. At each episode $k$, given an exploratory policy $\pi_k$, we build $M$ independent estimates $\{\widehat{Q}^{\pi_k, i}\}$ of the value function $Q^{\pi_k}$ by solving (11) on $M$ randomly generated subsets of the available trajectories in $D_k$. Then, we can leverage the disagreement among the models to approximate an optimistic or pessimistic estimate. An approximate pessimistic estimate can be obtained as:

$$\underline{Q}^\pi_1(s, a) = \widetilde{Q}^{\pi, i}_1(s, a) - \lambda \sqrt{\frac{1}{M} \sum_{i=1}^{M} \left( \widehat{Q}^{\pi, i}_1(s, a) - \widetilde{Q}^{\pi, i}_1(s, a) \right)^2}$$

where $\widetilde{Q}^{\pi, i}_1(s, a) = \frac{1}{M} \sum_{i=1}^{M} \widehat{Q}^{\pi, i}_1(s, a)$ is the mean value and $\lambda \in \mathbb{R}$ is a regularization coefficient. It is clear we can consider any learning algorithm in **B1** because the schema (**B2**, **B3**) only requires in input a target policy –selected by **B1**– and a set of trajectories. The final conservative condition is:

$$\sum_{t=1}^{T} \underline{Q}^{\pi_t}_1(s_1, \pi_t(s_1)) \geq (1 - \alpha_r \alpha) \sum_{t} Q^{\pi_b}_1(s_1, \pi_b(s_1)) \quad (13)$$

where $\alpha_r \in (0, 1)$ is used to tighten the conservative condition. This is to compensate the well-known over-estimation phenomenon arising with function approximation [e.g. 72, 73]. This means that the agent will aim to be more conservative than what really required.

---

[10]A major problem of IPS methods is that the importance weight is the product of density ratios over a trajectory, and tends to cause growth (even divergence) of the variance with the horizon $H$ [52, 53, 54]

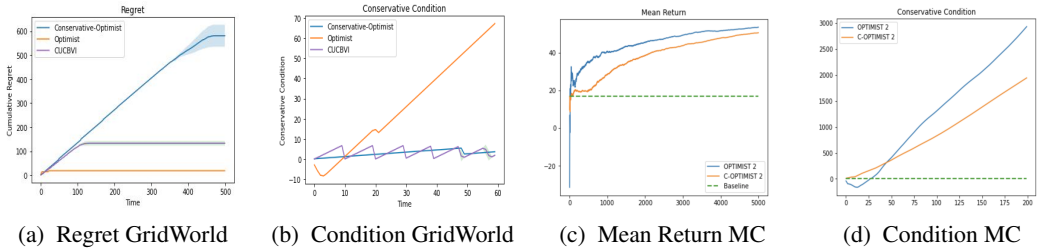

| (a) Regret GridWorld | (b) Condition GridWorld | (c) Mean Return MC | (d) Condition MC |

Figure 1: Experiments with COPTIMIST in a tabular MDP (Fig. 1a,1b) and with Continuous MountainCar (Fig. 1c,1d). The conservative condition reported in the figures is evaluated by running policy evaluation on the true MDP in the tabular case and by Monte-Carlo policy evaluation in Mountain Car. Regret is averaged over 5 runs.

## 5 Experiments

In this section, we evaluate the approaches discussed in Sec.3 and 4.

### 5.1 Evaluation of COPTIMIST

We first compare COPTIMIST to CUCBVI on a grid-word environment. CUCBVI is a model-based conservative exploration algorithm for tabular MDPs introduced in [9]. We also perform experiments on a continuous problem (MontainCar), and show that our algorithm COPTIMIST is also able to learn conservatively in such continuous environment contrary to CUCBVI. In both experiments, we consider the setting of parameter-based policy optimization with Gaussian hyper-policies.

**GridWorld.** We consider a discrete $3 \times 4$ grid-world with one goal state, one starting state and one trap state. The horizon is set to $H = 10$, the goal state gives a reward $0.5$, falling into a trap gives a reward $-1$ and the reward is $0$ for all the other states. We consider the space of policies such that $\pi(\pi^\star(s) \mid s) = p$ and $\pi(a \mid s) = 1 - p/A$ for $a \neq \pi^\star(s)$ where $\pi^\star$ is an optimal deterministic policy and $p = \frac{1}{1+e^{-\theta}}$ and $\theta \sim \nu_\xi = \mathcal{N}(\xi, 1)$. We also consider a uniform discretization of $[-5, 5]$ for $\xi$. We set the conservative level $\alpha = 0.1$ and choose the same baseline policy as in [9]. Finally, the regret is computed wrt. the best policy representable by COPTIMIST. Fig. 1b shows that both COPTIMIST and CUCBVI never violates the conservative condition. CUCBVI outperforms COPTIMIST in terms of regret but this is expected, since model-based algorithms usually performs better in tabular MDPs.

**Continuous MountainCar.** We then illustrate the conservative behavior of COPTIMIST, on the environment Continuous MountainCar [74] where previous conservative algorithm cannot be applied. The horizon $H$ is fixed to $300$. Our code is based on the original OPTIMIST implementation[11] where the policy has two learnable levels, one deterministic and the second a Gaussian hyperpolicy with a mean bounded in $[-1, 1] \times [0, 20]$ and a fixed co-variance matrix $\mathrm{diag}(0.15, 3)$. The baseline policy was extracted from OPTIMIST at mid-convergence, with $J(\pi_b)$ equals to $17$. We discretize the policy space as done in [11] with parameters $\kappa = 3$, $\delta = 0.2$, and the conservative level is chosen $\alpha = 0.5$. Theoretical confidence intervals may lead to over-pessimism, to avoid such a problem, we clipped the bonuses at well-chosen value of $20$.[12] In this experiment, we report the average reward instead of the regret since we don't have access to the exact optimal policy. As shown in Fig. 1c, COPTIMIST performs well even in continuous MDPs and almost on par with the non-conservative version. Furthermore, Fig. 1d shows that COPTIMIST never violates the conservative condition, even with the clipping of the confidence intervals. This shows that the original confidence intervals may be too large. While the original algorithm OPTIMIST converges quicker, it violates the safety constraints for a certain amount of time.

### 5.2 Extension to DeepRL

We evaluate the framework proposed in Sec. 4 on standard Gym/MuJoCo [75, 76] environments. We used Bootstrapped FQE and Ensemble-Based OPE with DQN and a variation of SAC.

---

[11] https://github.com/lorelupo/optimist

[12] See App. E for experiments with different clipping values, where the algorithm always converges to the optimal solution but requires more than $5,000$ iterations due to being overly conservative.

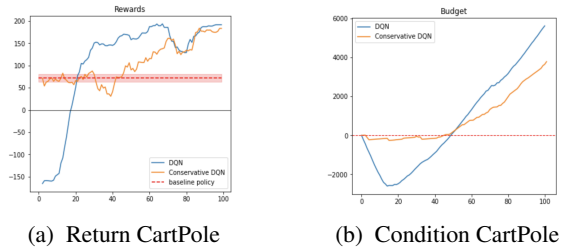

(a) Return CartPole    (b) Condition CartPole

Figure 2: Return (*Left*) and difference between return of the agent and the baseline policy (*Right*) for FQE-DQN. Results are averaged over 3 runs

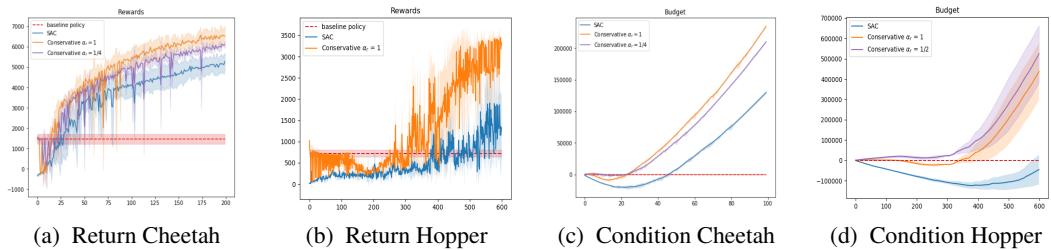

(a) Return Cheetah    (b) Return Hopper    (c) Condition Cheetah    (d) Condition Hopper

Figure 3: Return and conservative condition for Ensemble-SAC (Implementation based on SAC (https://github.com/vitchyr/rlkit) with the recommended hyper-parameters values. Results are averaged over 3 runs

**Conservative FQE-DQN.** We experiment a conservative version of DQN using Bootstrapped FQE as an OPE method on CartPole. Our implementation is based on [64, 65] and uses asymptotic confidence intervals to estimate the uncertainty of the evaluated policies. We used a single layer (64 hidden units) neural network agent. The baseline agent is obtained by stopping the training at mid-convergence. As shown in Fig. 2, even though confidence intervals provided by bootstrapping FQE are only valid asymptotically, the agent still acts conservatively and is able to converge to the maximum reward. It is interesting to observe that the loss in performance is small compared to the gain in safety. In fact, we can observe in Fig. 2b that non-conservative DQN largely violates the conservative condition at the beginning, while FQE-DQN is very close too the behavior of the baseline policy. This shows that OPE is viable method for conservative exploration in DRL, and does not seriously impact the overall performance.

**Conservative Ensemble-SAC.** OPE and related confidence estimations use ensemble learning with 5 evaluators. The regularization parameter $\lambda$ is set to $5$ in Cheetah and $1$ for Hopper. We use a looser $\lambda$ in Hopper since the algorithm requires more exploration to converge. The conservative level parameter is set to $\alpha = 0.2$ ; the baseline policy, including $\widehat{Q}^b$, is extracted from a mid-convergence run. Since $\widehat{Q}^b$ also suffers from uncertainties, it is estimated optimistically with the same $\lambda$. Despite looser theoretical guarantees, Conservative-SAC is able to sample safely most of the time, see Fig. 3. We highlight the ratio $\alpha_r$ is indeed required in the high dimensional environments to get a completely conservative agent as the pessimistic estimates also suffer from over-estimation. In these experiments, the baselines policies are safe and informative, they collect "good" samples. This provides a positive bias to the conservative algorithm that is able to reduce the need of exploration and leverage the relevant data to converge faster. This is quite an interesting aspect that requires further investigation both from a theoretical and empirical perspective.

## 6  Conclusion

We successfully applied conservative exploration to model free algorithms, where building counterfactual policy evaluation is challenging. We first investigated a theoretical approach using truncated Multi-Importance Sampling and derive the first regret bound for model-free conservative exploration. We then explored a generalisation to complex environments thanks to off-policy evaluation. An interesting direction for future work is to consider recent primal-dual techniques for OPE to design conservative algorithms, both from a theoretical and practical perspective.

## Acknowledgements

We particularly thank Matteo Pirotta and Alessandro Lazaric for their valuable insights, which significantly enriched the quality of our work. We also express our gratitude for their teachings in the Reinforcement Learning course as part of the Mathématiques, Vision, Apprentissage (MVA) master of Ecole Normale Superieure (ENS) Paris-Saclay.

## References

[1] Yishay Mansour, Aleksandrs Slivkins, and Vasilis Syrgkanis. Bayesian incentive-compatible bandit exploration. In *Proceedings of the Sixteenth ACM Conference on Economics and Computation*, pages 565–582. ACM, 2015.

[2] Yifan Wu, Roshan Shariff, Tor Lattimore, and Csaba Szepesvári. Conservative bandits. In *ICML*, volume 48 of *JMLR Workshop and Conference Proceedings*, pages 1254–1262. JMLR.org, 2016.

[3] Yihan Du, Siwei Wang, and Longbo Huang. A one-size-fits-all solution to conservative bandit problems. *CoRR*, abs/2012.07341, 2020.

[4] Abbas Kazerouni, Mohammad Ghavamzadeh, Yasin Abbasi, and Benjamin Van Roy. Conservative contextual linear bandits. In *NIPS*, pages 3910–3919, 2017.

[5] Evrard Garcelon, Mohammad Ghavamzadeh, Alessandro Lazaric, and Matteo Pirotta. Improved algorithms for conservative exploration in bandits. In *AAAI*, pages 3962–3969. AAAI Press, 2020.

[6] Sumeet Katariya, Branislav Kveton, Zheng Wen, and Vamsi K. Potluru. Conservative exploration using interleaving. In *AISTATS*, volume 89 of *Proceedings of Machine Learning Research*, pages 954–963. PMLR, 2019.

[7] Kun Wang, Canzhe Zhao, Shuai Li, and Shuo Shao. Conservative contextual combinatorial cascading bandit. *CoRR*, abs/2104.08615, 2021.

[8] Xiaojin Zhang, Shuai Li, and Weiwen Liu. Contextual combinatorial conservative bandits. *CoRR*, abs/1911.11337, 2019.

[9] Evrard Garcelon, Mohammad Ghavamzadeh, Alessandro Lazaric, and Matteo Pirotta. Conservative exploration in reinforcement learning. In *AISTATS*, volume 108 of *Proceedings of Machine Learning Research*, pages 1431–1441. PMLR, 2020.

[10] Eric Veach and Leonidas J. Guibas. Optimally combining sampling techniques for monte carlo rendering. In *SIGGRAPH*, pages 419–428. ACM, 1995.

[11] Matteo Papini, Alberto Maria Metelli, Lorenzo Lupo, and Marcello Restelli. Optimistic policy optimization via multiple importance sampling. In *ICML*, volume 97 of *Proceedings of Machine Learning Research*, pages 4989–4999. PMLR, 2019.

[12] L. Bottou, J. Peters, J. Quinonero-Candela, D. Charles, D. Chickering, E. Portugaly, D. Ray, P. Simard, and E. Snelson. Counterfactual reasoning and learning systems: The example of computational advertising. *Journal of Machine Learning Research*, 14:3207–3260, 2013.

[13] A. Swaminathan and T. Joachims. Counterfactual risk minimization: Learning from logged bandit feedback. In *Proceedings of The 32nd International Conference on Machine Learning*, 2015.

[14] G. Theocharous, P. Thomas, and M. Ghavamzadeh. Building personal ad recommendation systems for life-time value optimization with guarantees. In *Proceedings of the Twenty-Fourth International Joint Conference on Artificial Intelligence*, pages 1806–1812, 2015.

[15] P. Thomas, G. Theocharous, and M. Ghavamzadeh. High confidence off-policy evaluation. In *Proceedings of the Twenty-Ninth Conference on Artificial Intelligence*, 2015.

[16] P. Thomas, G. Theocharous, and M. Ghavamzadeh. High confidence policy improvement. In *Proceedings of the Thirty-Second International Conference on Machine Learning*, pages 2380–2388, 2015.

[17] M. Petrik, M. Ghavamzadeh, and Y. Chow. Safe policy improvement by minimizing robust baseline regret. In *Advances in Neural Information Processing Systems*, pages 2298–2306, 2016.

[18] Romain Laroche, Paul Trichelair, and Remi Tachet des Combes. Safe policy improvement with baseline bootstrapping. In *ICML*, volume 97 of *Proceedings of Machine Learning Research*, pages 3652–3661. PMLR, 2019.

[19] Thiago D. Simão and Matthijs T. J. Spaan. Safe policy improvement with baseline bootstrapping in factored environments. In *AAAI*, pages 4967–4974. AAAI Press, 2019.

[20] John Schulman, Sergey Levine, Pieter Abbeel, Michael I. Jordan, and Philipp Moritz. Trust region policy optimization. In *ICML*, volume 37 of *JMLR Workshop and Conference Proceedings*, pages 1889–1897. JMLR.org, 2015.

[21] Sham M. Kakade and John Langford. Approximately optimal approximate reinforcement learning. In *ICML*, pages 267–274. Morgan Kaufmann, 2002.

[22] Matteo Pirotta, Marcello Restelli, and Luca Bascetta. Adaptive step-size for policy gradient methods. In *NIPS*, pages 1394–1402, 2013.

[23] Felix Berkenkamp, Matteo Turchetta, Angela P. Schoellig, and Andreas Krause. Safe model-based reinforcement learning with stability guarantees. In *NIPS*, pages 908–918, 2017.

[24] Matteo Turchetta, Andrey Kolobov, Shital Shah, Andreas Krause, and Alekh Agarwal. Safe reinforcement learning via curriculum induction. In *NeurIPS*, 2020.

[25] Yinlam Chow, Ofir Nachum, Edgar A. Duéñez-Guzmán, and Mohammad Ghavamzadeh. A lyapunov-based approach to safe reinforcement learning. In *NeurIPS*, pages 8103–8112, 2018.

[26] Sanae Amani, Mahnoosh Alizadeh, and Christos Thrampoulidis. Linear stochastic bandits under safety constraints. In *NeurIPS*, pages 9252–9262, 2019.

[27] Aldo Pacchiano, Mohammad Ghavamzadeh, Peter L. Bartlett, and Heinrich Jiang. Stochastic bandits with linear constraints. In *AISTATS*, volume 130 of *Proceedings of Machine Learning Research*, pages 2827–2835. PMLR, 2021.

[28] Eitan Altman. *Constrained Markov decision processes*, volume 7. CRC Press, 1999.

[29] Liyuan Zheng and Lillian J. Ratliff. Constrained upper confidence reinforcement learning. In *L4DC*, volume 120 of *Proceedings of Machine Learning Research*, pages 620–629. PMLR, 2020.

[30] Yonathan Efroni, Shie Mannor, and Matteo Pirotta. Exploration-exploitation in constrained mdps. *CoRR*, abs/2003.02189, 2020.

[31] Dongsheng Ding, Kaiqing Zhang, Tamer Basar, and Mihailo R. Jovanovic. Natural policy gradient primal-dual method for constrained markov decision processes. In *NeurIPS*, 2020.

[32] Dongsheng Ding, Xiaohan Wei, Zhuoran Yang, Zhaoran Wang, and Mihailo R. Jovanovic. Provably efficient safe exploration via primal-dual policy optimization. In *AISTATS*, volume 130 of *Proceedings of Machine Learning Research*, pages 3304–3312. PMLR, 2021.

[33] Joshua Achiam, David Held, Aviv Tamar, and Pieter Abbeel. Constrained policy optimization. In *ICML*, volume 70 of *Proceedings of Machine Learning Research*, pages 22–31. PMLR, 2017.

[34] Chen Tessler, Daniel J. Mankowitz, and Shie Mannor. Reward constrained policy optimization. In *ICLR (Poster)*. OpenReview.net, 2019.

[35] Adam Stooke, Joshua Achiam, and Pieter Abbeel. Responsive safety in reinforcement learning by PID lagrangian methods. In *ICML*, volume 119 of *Proceedings of Machine Learning Research*, pages 9133–9143. PMLR, 2020.

[36] Homanga Bharadhwaj, Aviral Kumar, Nicholas Rhinehart, Sergey Levine, Florian Shkurti, and Animesh Garg. Conservative safety critics for exploration. *arXiv preprint arXiv:2010.14497*, 2020.

[37] Aviral Kumar, Aurick Zhou, George Tucker, and Sergey Levine. Conservative q-learning for offline reinforcement learning. *arXiv preprint arXiv:2006.04779*, 2020.

[38] Richard S. Sutton, David A. McAllester, Satinder P. Singh, and Yishay Mansour. Policy gradient methods for reinforcement learning with function approximation. In *NIPS*, pages 1057–1063. The MIT Press, 1999.

[39] Frank Sehnke, Christian Osendorfer, Thomas Rückstieß, Alex Graves, Jan Peters, and Jürgen Schmidhuber. Policy gradients with parameter-based exploration for control. In *ICANN (1)*, volume 5163 of *Lecture Notes in Computer Science*, pages 387–396. Springer, 2008.

[40] Marc Peter Deisenroth, Gerhard Neumann, and Jan Peters. A survey on policy search for robotics. *Found. Trends Robotics*, 2(1-2):1–142, 2013.

[41] Alberto Maria Metelli, Matteo Papini, Nico Montali, and Marcello Restelli. Importance sampling techniques for policy optimization. *J. Mach. Learn. Res.*, 21:141:1–141:75, 2020.

[42] Dongruo Zhou, Quanquan Gu, and Csaba Szepesvári. Nearly minimax optimal reinforcement learning for linear mixture markov decision processes. *CoRR*, abs/2012.08507, 2020.

[43] Chi Jin, Zeyuan Allen-Zhu, Sébastien Bubeck, and Michael I. Jordan. Is q-learning provably efficient? In *NeurIPS*, pages 4868–4878, 2018.

[44] Yash Chandak, Scott Niekum, Bruno Castro da Silva, Erik G. Learned-Miller, Emma Brunskill, and Philip S. Thomas. Universal off-policy evaluation. *CoRR*, abs/2104.12820, 2021.

[45] Yihao Feng, Ziyang Tang, Na Zhang, and Qiang Liu. Non-asymptotic confidence intervals of off-policy evaluation: Primal and dual bounds. *CoRR*, abs/2103.05741, 2021.

[46] Doina Precup, Richard S. Sutton, and Satinder P. Singh. Eligibility traces for off-policy policy evaluation. pages 759–766, 2000.

[47] Philip S. Thomas and Emma Brunskill. Data-efficient off-policy policy evaluation for reinforcement learning. In *ICML*, volume 48 of *JMLR Workshop and Conference Proceedings*, pages 2139–2148. JMLR.org, 2016.

[48] Scott Fujimoto, Herke van Hoof, and David Meger. Addressing function approximation error in actor-critic methods. In *ICML*, volume 80 of *Proceedings of Machine Learning Research*, pages 1582–1591. PMLR, 2018.

[49] Bo Dai, Ofir Nachum, Yinlam Chow, Lihong Li, Csaba Szepesvári, and Dale Schuurmans. Coindice: Off-policy confidence interval estimation. In *NeurIPS*, 2020.

[50] Yihao Feng, Ziyang Tang, Na Zhang, and Qiang Liu. Non-asymptotic confidence intervals of off-policy evaluation: Primal and dual bounds. *CoRR*, abs/2103.05741, 2021.

[51] Yihao Feng, Tongzheng Ren, Ziyang Tang, and Qiang Liu. Accountable off-policy evaluation with kernel bellman statistics. In *ICML*, volume 119 of *Proceedings of Machine Learning Research*, pages 3102–3111. PMLR, 2020.

[52] Yao Liu, Pierre-Luc Bacon, and Emma Brunskill. Understanding the curse of horizon in off-policy evaluation via conditional importance sampling, 2020.

[53] Qiang Liu, Lihong Li, Ziyang Tang, and Dengyong Zhou. Breaking the curse of horizon: Infinite-horizon off-policy estimation. In *NeurIPS*, pages 5361–5371, 2018.

[54] Nathan Kallus and Masatoshi Uehara. Efficiently breaking the curse of horizon: Double reinforcement learning in infinite-horizon processes. *CoRR*, abs/1909.05850, 2019.

[55] Cameron Voloshin, Hoang Minh Le, Nan Jiang, and Yisong Yue. Empirical study of off-policy policy evaluation for reinforcement learning. *CoRR*, abs/1911.06854, 2019.

[56] Ofir Nachum, Yinlam Chow, Bo Dai, and Lihong Li. Dualdice: Behavior-agnostic estimation of discounted stationary distribution corrections. In *NeurIPS*, pages 2315–2325, 2019.

[57] Ruiyi Zhang, Bo Dai, Lihong Li, and Dale Schuurmans. Gendice: Generalized offline estimation of stationary values. In *ICLR*. OpenReview.net, 2020.

[58] Shangtong Zhang, Bo Liu, and Shimon Whiteson. Gradientdice: Rethinking generalized offline estimation of stationary values. In *ICML*, volume 119 of *Proceedings of Machine Learning Research*, pages 11194–11203. PMLR, 2020.

[59] Hoang Minh Le, Cameron Voloshin, and Yisong Yue. Batch policy learning under constraints. In *ICML*, volume 97 of *Proceedings of Machine Learning Research*, pages 3703–3712. PMLR, 2019.

[60] Yaqi Duan, Zeyu Jia, and Mengdi Wang. Minimax-optimal off-policy evaluation with linear function approximation. In *ICML*, volume 119 of *Proceedings of Machine Learning Research*, pages 2701–2709. PMLR, 2020.

[61] Damien Ernst, Pierre Geurts, and Louis Wehenkel. Tree-based batch mode reinforcement learning. *J. Mach. Learn. Res.*, 6:503–556, 2005.

[62] Timothy P. Lillicrap, Jonathan J. Hunt, Alexander Pritzel, Nicolas Heess, Tom Erez, Yuval Tassa, David Silver, and Daan Wierstra. Continuous control with deep reinforcement learning. In *ICLR (Poster)*, 2016.

[63] Tuomas Haarnoja, Aurick Zhou, Pieter Abbeel, and Sergey Levine. Soft actor-critic: Off-policy maximum entropy deep reinforcement learning with a stochastic actor. In *ICML*, volume 80 of *Proceedings of Machine Learning Research*, pages 1856–1865. PMLR, 2018.

[64] Botao Hao, Xiang Ji, Yaqi Duan, Hao Lu, Csaba Szepesvári, and Mengdi Wang. Bootstrapping statistical inference for off-policy evaluation, 2021.

[65] Ilya Kostrikov and Ofir Nachum. Statistical bootstrapping for uncertainty estimation in off-policy evaluation, 2020.

[66] Marco A. Wiering and Hado van Hasselt. Ensemble algorithms in reinforcement learning. *IEEE Trans. Syst. Man Cybern. Part B*, 38(4):930–936, 2008.

[67] Kimin Lee, Michael Laskin, Aravind Srinivas, and Pieter Abbeel. SUNRISE: A simple unified framework for ensemble learning in deep reinforcement learning. *CoRR*, abs/2007.04938, 2020.

[68] Ian Osband, Charles Blundell, Alexander Pritzel, and Benjamin Van Roy. Deep exploration via bootstrapped DQN. In *NIPS*, pages 4026–4034, 2016.

[69] Richard Y. Chen, Szymon Sidor, Pieter Abbeel, and John Schulman. Ucb exploration via q-ensembles. *CoRR*, abs/1706.01502, 2017.

[70] Oron Anschel, Nir Baram, and Nahum Shimkin. Averaged-dqn: Variance reduction and stabilization for deep reinforcement learning. In *ICML*, volume 70 of *Proceedings of Machine Learning Research*, pages 176–185. PMLR, 2017.

[71] Rishabh Agarwal, Dale Schuurmans, and Mohammad Norouzi. An optimistic perspective on offline reinforcement learning. In *ICML*, volume 119 of *Proceedings of Machine Learning Research*, pages 104–114. PMLR, 2020.

[72] Hado van Hasselt. Double q-learning. In *NIPS*, pages 2613–2621. Curran Associates, Inc., 2010.

[73] Hado van Hasselt, Arthur Guez, and David Silver. Deep reinforcement learning with double q-learning. In *AAAI*, pages 2094–2100. AAAI Press, 2016.

[74] Andrew William Moore. Efficient memory-based learning for robot control. 1990.

[75] Greg Brockman, Vicki Cheung, Ludwig Pettersson, Jonas Schneider, John Schulman, Jie Tang, and Wojciech Zaremba. Openai gym, 2016.

[76] Emanuel Todorov, Tom Erez, and Yuval Tassa. Mujoco: A physics engine for model-based control. In *IROS*, pages 5026–5033. IEEE, 2012.

[77] Anna Harutyunyan, Marc G. Bellemare, Tom Stepleton, and Rémi Munos. Q($\lambda$) with off-policy corrections. In *ALT*, volume 9925 of *Lecture Notes in Computer Science*, pages 305–320, 2016.

[78] Nan Jiang and Lihong Li. Doubly robust off-policy value evaluation for reinforcement learning. In *ICML*, volume 48 of *JMLR Workshop and Conference Proceedings*, pages 652–661. JMLR.org, 2016.

[79] Mehrdad Farajtabar, Yinlam Chow, and Mohammad Ghavamzadeh. More robust doubly robust off-policy evaluation. In *ICML*, volume 80 of *Proceedings of Machine Learning Research*, pages 1446–1455. PMLR, 2018.

[80] Miroslav Dudík, John Langford, and Lihong Li. Doubly robust policy evaluation and learning. In *ICML*, pages 1097–1104. Omnipress, 2011.

[81] Volodymyr Mnih, Koray Kavukcuoglu, David Silver, Alex Graves, Ioannis Antonoglou, Daan Wierstra, and Martin A. Riedmiller. Playing atari with deep reinforcement learning. *CoRR*, abs/1312.5602, 2013.

[82] Matteo Hessel, Joseph Modayil, Hado van Hasselt, Tom Schaul, Georg Ostrovski, Will Dabney, Dan Horgan, Bilal Piot, Mohammad Gheshlaghi Azar, and David Silver. Rainbow: Combining improvements in deep reinforcement learning. In *AAAI*, pages 3215–3222. AAAI Press, 2018.

# Appendix

## Table of Contents

## A  Off-Policy Policy Evaluation

Following [55], we can group these methods into three categories: inverse propensity score (IPS), direct and hybrid methods. IPS methods [46] rely on some form of importance sampling, e.g., per-decision Importance sampling, Weighted Importance Sampling. The idea is to weight rewards by an importance ratio between the policy to be evaluated and the policy (or policies) used to collect the samples. These methods may lead both to biased or unbiased estimates of the policy performance. Direct methods [59, 77] aim to directly estimate the value function of the evaluation policy. These methods may be both model-free and model-based. These methods can be easily extended to function approximation since the OPE problem is often reduced to a regression problem. Finally, hybrid methods [78, 47, 79] combine an IPS and a direct approach. A classical example of hybrid method is the doubly robust estimator [80]. Refer to [55] for a broader review and empirical comparison.

Note that for IPS, it is possible to build non-asymptotic confidence bounds [15, 47, 41]. Unfortunately, a major problem of IPS methods is that the importance weight is the product of density ratios over a trajectory, and tends to cause growth (even divergence) of the variance with the horizon $H$. Despite improvements can be obtained using advanced techniques (e.g., self-normalized, balance heuristics, control variates), the curse of dimensionality remains one of the limitations of IPS techniques. This and the fact that it is often necessary to retain the behavioral policies, make IPS methods not well suited for large DRL approaches.

On the other hand, for direct methods, under certain condition on the data generating process and function approximation, it is also possible to build confidence intervals around the OPE estimate [e.g., 49, 45]. Direct methods approaches have become more popular in the recent years due to their scalability to function approximation.

## B  Concentration Inequality for the Robust Balance Heuristic Estimate

For $\alpha \in [0, \infty]$, $d_\alpha$ is the exponentiated $\alpha$-Rényi divergence, measuring the divergence between two probability measures $P$ and $Q$, defined as:

$$d_\alpha(P||Q) = \exp\left\{\frac{1}{1-\alpha}\log\int_{\mathcal{Z}}(w_{P/Q})^\alpha dQ\right\}$$

And $w_{P/Q}$ is the Radon-Nikodym derivative of P wrt. Q.

We report the concentration inequality using RBH introduced in [11].

**Theorem B.1.** *Let $P$ and $\{Q_m\}_{m=1}^M$ be probability measures on the measurable space $\{\mathcal{Z}, \mathcal{F}\}$ such that $P << Q_m$ and there exists $\epsilon \in (0, 1]$ s.t. $d_{1+\epsilon}(P||Q_m) < \infty$ for $m = 1, ..., M$. Let $f : \mathcal{Z} \to \mathbb{R}_+$ be a bounded non-negative function, i.e., $||f||_\infty < \infty$. Let $J_k^{RBH}(\pi)$ be the truncated balance heuristic estimator of f, as defined in 3, using $N_m$ i.i.d. samples from each $Q_m$.*

*Let $C_k = \left(\frac{(k-1)d_{1+\epsilon}(P||\Phi)^\epsilon}{\log \frac{1}{\delta}}\right)^{\frac{\epsilon}{1+\epsilon}}$ be the adaptive threshold of the truncated weights, then with probability at least $1-\delta$:*

$$J_k^{RBH}(\pi) \leq J(\pi) + ||f||_\infty \left(\sqrt{2} + \frac{1}{3}\right) \left(\frac{d_{1+\epsilon}(P||\Phi)\log\frac{1}{\delta}}{k-1}\right)^{\frac{\epsilon}{1+\epsilon}}$$

*And with probability at least $1-\delta$:*

$$J_k^{RBH}(\pi) \geq J(\pi) - ||f||_\infty \left(\sqrt{2} + \frac{4}{3}\right) \left(\frac{d_{1+\epsilon}(P||\Phi)\log\frac{1}{\delta}}{k-1}\right)^{\frac{\epsilon}{1+\epsilon}}$$

*Where: $\Phi = \sum_{m=1}^M Q_m$ is a finite mixture and $\sum_{m=1}^M N_m = k-1$.*

To simplify the notations we set: $\beta_k(\pi) = ||f||_\infty \left(\sqrt{2} + \frac{4}{3}\right) \left(\frac{d_{1+\epsilon}(P||\Phi)\log\frac{1}{\delta}}{k-1}\right)^{\frac{\epsilon}{1+\epsilon}}$ and use equivalently to theorem B.1:

$$|J_k^{\text{RBH}}(\pi) - J(\pi)| \leq \beta_k(\pi) = O\left((k-1)^{-\frac{\epsilon}{1+\epsilon}}\right). \tag{14}$$

## C  Algorithms additional explanations

### C.1  Conservative OPTIMIST

We report the pseudo-code of COPTIMIST defined in section 3.

---
**Algorithm 3** COPTIMIST

**Input:** baseline policy $\pi_b$
**for** $k = 1, \ldots$ **do**
　　Select policy $\pi_k = \text{argmax}_{\pi \in \Pi}\{J_k^{\text{RBH}}(\pi) + \beta_k(\pi)\}$
　　**if** (6) not verified **then**
　　　　Set $\pi_k = \pi_b$
　　**end if**
　　Run policy $\pi_k$ and observe trajectory $\tau_k = (s_{kh}, a_{kh}, r_{kh}, s_{k,h+1})_{h \in [H]}$
　　Set $D_{k+1} = D_k \cup \{\tau_k\}$
**end for**

---

The optimization step $\text{argmax}_{\pi \in \Pi}\{J_k^{\text{RBH}}(\pi) + \beta_k(\pi)\}$ can be intractable, especially when $\Pi$ is compact. Instead of optimizing on the entire set $\Pi$, we find the maximum of a progressively finer grid $\tilde{\Pi}_k$ with $\lceil \varkappa_k \rceil^d$ vertices. The conservative algorithm is called COPTIMIST 2:

---
**Algorithm 4** COPTIMIST 2

**Input:** baseline policy $\pi_b$
**for** $k = 1, \ldots$ **do**
　　Discretize $\Pi$ with a uniform grid $\tilde{\Pi}$ of $\lceil \varkappa_k \rceil^d$ points
　　Select policy $\pi_k = \text{argmax}_{\pi \in \Pi}\{J_k^{\text{RBH}}(\pi) + \beta_k(\pi)\}$
　　**if** (6) not verified **then**
　　　　Set $\pi_k = \pi_b$
　　**end if**
　　Run policy $\pi_k$ and observe trajectory $\tau_k = (s_{kh}, a_{kh}, r_{kh}, s_{k,h+1})_{h \in [H]}$
　　Set $D_{k+1} = D_k \cup \{\tau_k\}$
**end for**

---

We can prove a similar regret bound for COPTIMIST 2. See appendix D.3.

## C.2 Conservative DQN

*Deep-Q-Learning* (DQN) [81] is the state-of-the-art RL agent for environments with discrete agents. It directly learns the optimal $Q$-function by minimizing the optimal Bellman residual as follows:

$$\widehat{Q}_h^\star = \min_{g \in \mathcal{G}^\star} \frac{1}{N} \sum_{i=1}^n \left( g(s_h^i, a_h^i) - r_h^i - \max_{a' \in \mathcal{A}} \mathbb{E}\big[\widehat{Q}_{h+1}(s_{h+1}^i, a')\big] \right)^2 \quad (15)$$

$\mathcal{G}^\star$ is parametrized by a Neural Network in the DQN agent. An important number of tricks have been introduced to enhance and stabilize learning, combined in [82]. The DQN agent usually follows and $\epsilon$-greedy policy with respect to the learned optimal $Q$-values, with $\epsilon$ decreasing over episodes.

---

**Algorithm 5** Conservative DQN

1: **Input:** baseline policy $\pi_b$, epsilon schedule $(\epsilon_k)_k$, performance of the baseline $J(\pi_b) = J_b$, conservative parameter $\alpha$
2: **Initialize:** replay memory $D$ with transitions sampled with $\pi_b$, action-value function $Q$ with random weights
3: **for** $k = 1, \ldots$ **do**
4:     Let $\pi_k$ the $\epsilon_k$-greedy policy of $Q$
5:     **Eval** $J^{pessimistic}(\pi_k)$ with 6
6:     **if** $\sum_{j=1}^k J^{pessimistic}(\pi_j) < k(1-\alpha)J(\pi_b)$ : **then**
7:         Set $\pi_k = \pi_b$
8:     **end if**
9:     Run policy $\pi_k$ and observe trajectory $\tau_k = (s_{kh}, a_{kh}, r_{kh})_{h \in [H]}$
10:    Set $D = D \cup \{\tau_k\}$
11:    Update the action value function $Q$ minimizing (15)
12: **end for**

---

**Algorithm 6** Bootstrapping Fitted Q-Evaluation

1: **Input:** baseline policy $\pi_b$, $\epsilon_k$-greedy policy $\pi_k$, replay memory $D$ of length $n$, number of bootstrapping estimates $M$
2: **Compute** $\hat{Q}_{\pi_k}$ with fitted Q-iteration (FQI) using all samples from $D$, and **estimate** $\hat{v}_{\pi_k} \approx \frac{1}{N_s} \sum_{s \sim d_0} \hat{Q}_{\pi_k}(s, \pi_k(s))$
3: **for** $m = 1, \ldots, M$ **do**
4:     Form $D_m$ by sampling $n$ transitions from $D$ with replacement
5:     Similarly, estimate $\hat{v}_{\pi_k}^m$ from $D_m$ using FQI
6: **end for**
7: **Compute** $q_{1-\delta/2}^{\pi_k}$, the $1 - \delta/2$ quantile of $\{\hat{v}_{\pi_k}^m - \hat{v}_{\pi_k}\}_{m=1}^M$
8: **Return** $J^{pessimistic}(\pi_k) = \hat{v}_{\pi_k} - q_{1-\delta/2}^{\pi_k}$

---

## C.3 Conservative Ensemble SAC

### C.3.1 Soft Actor Critic (SAC)

*Soft Actor Critic* (SAC) [63] is an efficient RL algorithm for continuous control problems.

It estimates the value function of a sampling policy and maximizes the *entropy*-regularized average return $\mathbb{E}\left[\sum_h r(s_h, a_h) + \eta \mathcal{H}(\pi(\cdot|s_h)) | a_h \sim \pi(\cdot|s_h)\right]$ with the entropy $\mathcal{H}$. The algorithm alternate between collecting samples with the sampling policy and updating its value function.

$$\widehat{Q}^{k+1} = \min_{g \in \mathcal{G}} \frac{1}{N} \sum_{i=1}^{N} L_Q(s_h^i, a_h^i, s_{h+1}^i)$$

$$\widehat{\pi}^{k+1} = \min_{\pi \in \mathcal{F}} \frac{1}{N} \sum_{i=1}^{N} L_\pi(s_h^i, a_h^i, s_{h+1}^i)$$

with

$$L_Q(s, a, s') = r(s,a) + \gamma \mathbb{E}_{a' \sim \widehat{\pi}^k} \big[\widehat{Q}^k(s', a') - \eta \log \widehat{\pi}^k(a'|s')\big] - g(s, a)\Big)^2 \quad (16)$$

$$L_\pi(s) = -\mathbb{E}_{a' \sim \pi}\big[\widehat{Q}_{k+1}(s, a') - \eta \log \pi(a'|s)\big] \quad (17)$$

and $\mathcal{F}$ and $\mathcal{G}$ are two arbitrary function space, usually parametrized by Neural Network weights.

**Weighted Bellman Backup** Additional tricks inspired from [67] might be added to prevent overestimations of the value function. In SAC, the estimated Q-values are trained with bootstrapped targets which can be erroneous, especially at the beginning of learning. The errors back-propagates towards the estimators, and might prevent convergence. To cope with it, the loss is multiplied by an additional term $\mathcal{L}_{wbb}$ controlled by a parameter $T > 0$. It penalizes samples with high variances that can harm the training process. The penalty term is defined as:

$$\mathcal{L}_{wbb}(s, a) = \sigma\left(-Q_{std}(s, a) T\right) + 0.5$$

The new loss for the $Q$-values is then:

$$\mathcal{L}_{WQ} = \frac{1}{N} \sum_{i=1}^{N} L_Q(s_h^i, a_h^i, s_{h+1}^i) \mathbb{E}_{a' \sim \widehat{\pi}^k}\left[\mathcal{L}_{wbb}(s_{h+1}^i, a')\right] \quad (18)$$

### C.3.2 Conservative Ensemble SAC

Conservative Ensemble SAC leverages Ensemble Learning to estimate the value function of the sampling policy before collecting one episode of data, see equation 13 of section 3.

---
**Algorithm 7** Conservative Ensemble SAC
---
1: **Input:** baseline policy $\pi_b$
2: **for** $k = 1, \ldots$ **do**
3:     **if** 13 not verified: **then**
4:         Set $\pi_k = \pi_b$
5:     **end if**
6:     Collect one trajectory $\tau_k$ using $\pi_k$.
7:     // TRAIN
8:     Update $\widehat{Q}_k$ and $\hat{\pi}_k$ with (16 or 18) and 17
9: **end for**
---

## D  Regret Analysis of COPTIMIST

In order to prove our main theorem, we first define a confidence schedule such that our regret bound holds with probability at least $1 - \delta$. Then, we demonstrate that the last episode played conservatively $K_\epsilon$ exists and is bounded. Finally, we prove a regret bound of COPTIMIST in theorem 3.1.

## D.1 Confidence bound analysis

The goal of this Lemma is two-fold. First, we want to make sure the RBH estimate $J^{\text{RBH}}(\pi_k)$ stays in its confidence intervals during all the episodes. We also need to make sure the optimal policy remain in its confidence intervals with high probability. Considering $\Pi$ may be compact, there may be more than one optimal policy. Thus, we introduce a discretization of the policy space $\Pi$, in order to prove, for all episodes and *all vertices* of $\tilde{\Pi}_k$:

$$J(\pi^\star) - J_k^{\text{RBH}}([\pi^\star]_k) \leq \beta_k([\pi^\star]_k) + \frac{LDd}{\varkappa_k}$$

Note this discretization is done only for the purposes of the proof and is not present in the algorithm.

**Lemma D.1.** *Let $\tilde{\Pi}$ be a mesh of $\Pi$ with discretization size $\varkappa$ and $[\pi] \in \tilde{\Pi}$ the closest point from $\pi \in \Pi$. Moreover, for all $\pi \in \Pi$, let define:*

$$\tilde{\mathcal{E}} = \bigcap_{k=1}^{K} \bigcap_{\lceil \varkappa_k \rceil^d} \left\{ J(\pi^\star) - J_k^{RBH}([\pi^\star]_k) \leq \beta_k([\pi^\star]_k) + \frac{LDd}{\varkappa_k} \right\}$$

$$\mathcal{E}^+ = \bigcap_{k=1}^{K} \left\{ J(\pi_k) - J_k^{RBH}(\pi_k) \leq \beta_k(\pi_k) \right\}$$

$$\mathcal{E}^- = \bigcap_{k=1}^{K} \left\{ J_k^{RBH}(\pi_k) - J(\pi_k) \leq \beta_k(\pi_k) \right\}$$

$$\mathcal{E} = \tilde{\mathcal{E}} \bigcap \mathcal{E}^+ \bigcap \mathcal{E}^-$$

*Under discretization schedule $\varkappa_k = dk^2$ and confidence schedule $\delta_k = \frac{6\delta}{k^2 \pi^2 (2+d^d k^{2d})}$:*

$$\mathbf{P}(\mathcal{E}) \geq 1 - \delta$$

*Proof.* For all $\pi \in \Pi$, for all $k = 1, ..., K$

$$J(\pi^\star) - J_k^{RBH}([\pi^\star]) \leq |J(\pi^\star) - J([\pi^\star]_k))| + J([\pi^\star]_k)) - J_k^{RBH}([\pi^\star]_k)$$
$$\leq \frac{LDd}{\varkappa_k} + \beta_k([\pi^\star]_k)$$

Performing a union over the $\lceil \varkappa_k \rceil^d$ vertices of $\Pi$ and subsequent steps over $k = 1, ..., K$, we have, with confidence $\delta' = \sum_{k=1}^{K} \varkappa_k^d \delta_k$:

$$\mathbf{P}(\tilde{\mathcal{E}}) \leq 1 - \delta'$$

Moreover we have with confidence schedule $\delta'' = \sum_{k=1}^{K} \delta_k$:

$$\mathbf{P}(\mathcal{E}^+) = \mathbf{P}(\mathcal{E}_t^-) \leq 1 - \delta''$$

Finally, performing the union bound over these events, we have with confidence schedule $\delta' + 2\delta'' = \sum_{k=1}^{K} (2 + \varkappa_k^d) \delta_k$:

$$\mathbf{P}(\mathcal{E}) \leq 1 - \delta' - 2\delta''$$

Taking a discretization schedule $\varkappa_k = k^2 d$, using confidence schedule $\delta_k = \frac{6\delta}{k^2 \pi^2 (2+d^d k^{2d})}$ we can prove that:

$$\mathbf{P}(\mathcal{E}) \leq 1 - \delta$$

□

**Discrete setting** Note the discretization is no longer required when $\Pi$ is discrete, removing the need of $\tilde{\mathcal{E}}$. The discretization schedule becomes $\delta_t = \frac{3\delta}{t^2\pi^2|\Pi|}$

## D.2 Proof of theorem 3.1

We begin to prove the following lemma:

**Lemma D.2.** *Under event $\mathcal{E}$, the last episode $K_\epsilon < \infty$ played conservatively exists.*

*Proof.* We analyze when the slack accumulated at episode k become positive for all subsequent steps. We then deduce that (2) will hold for following episodes and the algorithm stops to play conservatively. At episode $k$, the slack is positive when this condition meets:

$$J_k^{\text{RBH}}(\pi_k) - \beta_k(\pi_k) \geq (1-\alpha)J(\pi_b) \tag{19}$$

$$J_k^{\text{RBH}}(\pi_k) - J(\pi^\star) + J(\pi^\star) - J(\pi_b) + \alpha J(\pi_b) \geq \beta_k(\pi_k) \tag{20}$$

$$J_k^{\text{RBH}}(\pi_k) - J(\pi^\star) + \Delta_b + \alpha J(\pi_b) \geq \beta_k(\pi_k) \tag{21}$$

$$J_k^{\text{RBH}}(\pi_k) - J(\pi_k) + J(\pi_k) - J(\pi^\star) + \Delta_b + \alpha J(\pi_b) \geq \beta_k(\pi_k) \tag{22}$$

Under event $\mathcal{E}$, we have the following inequalities:

$$J_k^{\text{RBH}}(\pi_k) - J(\pi_k) \geq -\beta_k(\pi_k) \tag{23}$$

And:

$$J(\pi_k) - J(\pi^\star) = J(\pi_k) - J_k^{\text{RBH}}([\pi^\star]_k) + J_k^{\text{RBH}}([\pi^\star]_k) - J(\pi^\star) \tag{24}$$

$$\geq J(\pi_k) - J_k^{\text{RBH}}([\pi^\star]_k) - \beta_k([\pi^\star]_k) - \frac{LDd}{\varkappa_k} \tag{25}$$

$$\geq J(\pi_k) - J_k^{\text{RBH}}(\pi_k) - \beta_k^\epsilon(\pi_k) - \frac{LDd}{\varkappa_k} \tag{26}$$

$$\geq -2\beta_k(\pi_k) - \frac{LDd}{\varkappa_k} \tag{27}$$

In the remaining part of the proof we use a discretization schedule $\varkappa_k = dk^2$. Note this has no impact whatsoever on the algorithm, as the discretization is only hypothetical. We can now lower bound the LHS and upper bound the RHS of 22 to have a sufficient condition for all subsequent steps:

$$-\beta_k(\pi_k) - 2\beta_k(\pi_k) + \Delta_b + \alpha J(\pi_b) - \frac{LD}{k^2} \geq \beta_k(\pi_k) \tag{28}$$

$$\frac{k\left(\Delta_b + \alpha J(\pi_b) - \frac{LD}{k^2}\right)^{\frac{\epsilon}{1+\epsilon}}}{\log\left(\frac{1}{\delta_k}\right)} \geq 4v_\epsilon \left(\sqrt{2} + \frac{4}{3}\right)^{\frac{\epsilon}{1+\epsilon}} \tag{29}$$

$K_\epsilon$ exists since $\lim_{k \to \infty} \frac{k\left(\Delta_b + \alpha J(\pi_b) - \frac{LD}{k^2}\right)^{\frac{\epsilon}{1+\epsilon}}}{\log\left(\frac{1}{\delta_k}\right)} = \infty$.

$\square$

**Remark** Note an upper bound in $O(\log K)$ can be retrieved for $K_\epsilon$.

**Discrete setting** Note the term $\frac{LD}{K_m^2}$ is removed when the set of available policies is discrete.

### D.2.1 Technical Lemmas

Before diving into the complete proof of our theorem, let's state two technical Lemmas.

**Lemma D.3.** *For all $\alpha \in (0,1)$:*

$$\sum_{t=1}^{T} t^{-\alpha} \leq \frac{T^{1-\alpha}}{1-\alpha}$$

*Proof.*

$$\sum_{t=1}^{T} t^{-\alpha} \leq \int_{1}^{T+1} t^{-\alpha}\, dt$$
$$= \frac{1}{1-\alpha}\left((T+1)^{1-\alpha} - 1\right)$$
$$\leq \frac{T^{1-\alpha}}{1-\alpha}$$

□

**Lemma D.4.** *For any $m \in \mathbb{R}_+^\star$, $\alpha \in (0,1)$ and any $c_1, c_2 > 0$:*

$$-c_1 m + c_2 m^{1-\alpha} \leq \frac{\alpha(1-\alpha)^{1/\alpha - 1}}{1-\alpha} \frac{c_2^{1/\alpha}}{c_1^{1/\alpha - 1}}$$

*Proof.* Let $f(m) = -c_1 m + c_2 m^{1-\alpha}$. $f \in \mathcal{C}_2(\mathbb{R}_\star^+)$, and:

$$f'(x) = -c_1 + c_2(1-\alpha)x^{-\alpha}, \qquad f''(x) = -c_2 \alpha (1-\alpha) x^{-(\alpha+1)}$$

Since $\alpha \in (0,1)$, $f$ is concave and its maximum $x^\star$ is such that $f'(x^\star) = 0$ that is to say:

$$\frac{c_2(1-\alpha)}{c_1} = (x^\star)^\alpha$$

Hence $x^\star = \left(\frac{c_2(1-\alpha)}{c_1}\right)^{1/\alpha}$ and $\max_{x \in \mathbb{R}_+^\star} f(x) = f(x^\star) = \frac{\alpha(1-\alpha)^{1/\alpha - 1}}{1-\alpha} \frac{c_2^{1/\alpha}}{c_1^{1/\alpha - 1}}$

□

Here is the total formula of our theorem:

### D.2.2 Theorem proof

In this subsection, we denote $\mathbb{C} = \left(\sqrt{2} + \frac{4}{3}\right)$.

**Theorem D.5.** *Under Asm. 1, 2 and 3, for any $K$ and conservative level $\alpha$, there exists a numerical constant $\beta > 0$ such that the regret of* COPTIMIST *is bounded as*

$$R(K) \leq \beta R_{\text{OPTIMIST}}(K | \Lambda_k) + \beta \frac{\Delta_b}{\alpha J(\pi_b)} \Big( \underbrace{\varepsilon(1+\varepsilon) \frac{(\nu_\varepsilon L_K(\delta))^{1+1/\varepsilon}}{(\Delta_b + \alpha \mu_b)^{1/\varepsilon}} + H}_{:= \text{Ⓑ}} + \underbrace{\frac{LD\pi^2}{6}}_{\text{Ⓐ}} \Big) \quad (30)$$

*and $\sum_{l=1}^{k} J(\pi_l) \geq (1-\alpha)k J(\pi_b)$ is verified at every $k \in [K]$ with probability $1-\delta$. Note that $L_K(\delta)$ is a logarithmic term in $K$. Under Asm. 3 we have that $L_K(\delta) = \log\left(\frac{\pi^2 K^2}{6\delta}\right) + d\log\left(1 + dK^2\right)$.*

*Proof.* Rewriting the regret:

$$R(K) = |\Lambda_K^c| \Delta_b + R_{OPTIMIST}(K, |\Lambda_K|)$$

Regarding the first term, the only difference with the original proof of Papini & al. [11] is that the baseline policy can be played instead of the optimistic one which would only tighten the expectation $\mu_t$. Thus, it can be bounded with $\beta R_{OPTIMIST}(T)$ where $\beta$ is a strictly positive constant. Only the bound $|\Lambda_T^c|$ remains.

Let $K_\epsilon$ be the last episode played conservatively. Its existence and upper bound is guaranteed by Lemma D.2. Before $K_\epsilon$, the condition (2) is not verified, so:

$$\sum_{l \in \Lambda_{K_\epsilon-1} \cup \{K_\epsilon\}} \left(J_l^{\text{RBH}}(\pi_l) - \beta_l(\pi_l)\right) + \sum_{l \in \Lambda_{K_\epsilon-1}^c} J(\pi_b) \leq (1-\alpha) \sum_{l=1}^{K_\epsilon} J(\pi_b)$$

$$\alpha K_\epsilon J(\pi_b) \leq \sum_{l \in \Lambda_{K_\epsilon-1} \cup \{K_\epsilon\}} \left(J(\pi_b) - J_l^{\text{RBH}}(\pi_l) + \beta_l(\pi_l)\right)$$

In the remaining part of the proof we use a discretization schedule $\varkappa_k = dk^2$. Note this has no impact whatsoever on the algorithm, as the discretization is only hypothetical.

Let's bound $\Delta_l = J(\pi_b) - J_l^{\text{RBH}}(\pi_l) + \beta_l(\pi_l)$ over the non-conservative episodes:

$$\Delta_l = J(\pi_b) - J_l^{\text{RBH}}(\pi_l) + \beta_l(\pi_l) \tag{31}$$

$$= J(\pi_b) - J(\pi^\star) + J(\pi^\star) - J_l^{\text{RBH}}(\pi_l) + \beta_l(\pi_l) + \frac{LD}{l^2} \tag{32}$$

$$= J(\pi_b) - J(\pi^\star) + J(\pi^\star) - J([\pi^\star]_l) + J([\pi^\star]_l) - J_l^{\text{RBH}}(\pi_l) + \beta_l(\pi_l) + \frac{LD}{l^2} \tag{33}$$

$$\leq -\Delta_b + J_l^{\text{RBH}}(\pi_l) + \beta_l(\pi_l) - J_l^{\text{RBH}}(\pi_l) + \beta_l(\pi_l) + \frac{LD}{l^2} \tag{34}$$

$$\leq -\Delta_b + 2\mathbb{C}\left(\frac{v_\epsilon \log\left(\frac{1}{\delta_l}\right)}{l}\right)^{\frac{\epsilon}{1+\epsilon}} + \frac{LD}{l^2} \tag{35}$$

For episode $K_\epsilon$, the value difference is bounded by $H$.

Now, let's upper-bound $\sum_{l \in \{\Lambda_{K_\epsilon-1}\}} \left(\frac{1}{l}\right)^{\frac{\epsilon}{1+\epsilon}}$. In the worst case scenario, $\Lambda_{K_\epsilon-1} = \{1, 2, \ldots, |\Lambda_{K_\epsilon-1}|\}$. Lemma D.3 gives:

$$\sum_{l \in \{\Lambda_{K_\epsilon-1}\}} \left(\frac{1}{l}\right)^{\frac{\epsilon}{1+\epsilon}} \leq \sum_{l=1}^{|\Lambda_{K_\epsilon-1}|} \left(\frac{1}{l}\right)^{\frac{\epsilon}{1+\epsilon}} \leq \frac{|\Lambda_{K_\epsilon-1}|^{1-\frac{\epsilon}{1+\epsilon}}}{1 - \frac{\epsilon}{1+\epsilon}}$$

Besides:

$$\sum_{l \in \{\Lambda_{K_\epsilon-1}\}} \frac{LD}{l^2} \leq \sum_{l=1}^{|\Lambda_{K_\epsilon-1}|} \frac{LD}{l^2} \leq \frac{LD\pi^2}{6}$$

In addition, the term $\log \frac{1}{\delta_l}$ is bounded with $\log \frac{1}{\delta_K}$.

Considering $\alpha K_\epsilon J(\pi_b) = \alpha \left(|\Lambda_{K_\epsilon-1}| + |\Lambda_{K_\epsilon-1}^c| + 1\right) J(\pi_b)$, we have:

$$\alpha(|\Lambda_{K_\epsilon-1}^c| + 1)J(\pi_b) \leq -(\Delta_b + \alpha J(\pi_b))|\Lambda_{K_\epsilon-1}| + 2\mathbb{C}\left(v_\epsilon \log\left(\frac{1}{\delta_K}\right)\right)^{\frac{\epsilon}{1+\epsilon}} \frac{|\Lambda_{K_\epsilon-1}|^{1-\frac{\epsilon}{1+\epsilon}}}{1 - \frac{\epsilon}{1+\epsilon}} + H + \frac{LD\pi^2}{6}$$

Applying Lemma D.4, noting that $|\Lambda_{K_\epsilon-1}^c| + 1$ is exactly $|\Lambda_T^c|$ and including the term $\mathbb{C}$ in $\beta$ concludes the proof.

$\square$

**Remark** Using $\delta_{K_\epsilon}$ instead of $\delta_K$ allows having an upper bound in $\log(\log K)$ instead of $\log K$.

## D.3 COPTIMIST 2, regret analysis.

The regret analysis of COPTIMIST 2 is very similar to COPTIMIST, except from the fact that the discretization is actually performed by the algorithm. We use a different discretization schedule: $\varkappa_k = k^{\frac{1}{\kappa}}, \kappa > 2$.

**Theorem D.6.** *Under Asm. 1, 2 and 3, using discretization schedule $\varkappa_k = k^{\frac{1}{\kappa}}$, for any $K$ and conservative level $\alpha$, there exists a numerical constant $\beta > 0$ such that the regret of COPTIMIST is bounded as*

$$R(K) \leq \beta R_{\text{OPTIMIST2}}(K|\Lambda_k) + \beta \frac{\Delta_b}{\alpha J(\pi_b)} \left( \varepsilon(1+\varepsilon) \frac{(\nu_\varepsilon L_K^2(\delta))^{1+1/\varepsilon}}{(\Delta_b + \alpha \mu_b)^{1/\varepsilon}} + H + \frac{\kappa}{\kappa - 1} dLDK^{1-\frac{1}{\kappa}} \right) \quad (36)$$

*and $\sum_{l=1}^k J(\pi_l) \geq (1-\alpha)kJ(\pi_b)$ is verified at every $k \in [K]$ with probability $1-\delta$. Note that $L_K(\delta)$ is a logarithmic term in $K$. Under Asm. 3 we have that $L_K^2(\delta) = \log\left(\frac{\pi^2 K^2}{6\delta}\right) + d\log\left(1 + K^{\frac{1}{\kappa}}\right)$.*

For further details on $R_{\text{OPTIMIST2}}$, see theorem 4 of [11]

## E Experiments Details

### E.1 Grid World

We can illustrate the dynamic of COPTIMIST by plotting the evolution of the estimator $J_k^{RBH}$ for the top 5 policies and their uncertainty gap during execution for $\sigma = 1$, slack $\alpha = 0.1$ and $T = 550$. The number of play of each policy is given in array 5. In figure 4 we see that the uncertainty shrink for policies being often played (e.g. policy 1) as well as for correlated policies (e.g. policy 2) contrary to policy 5.

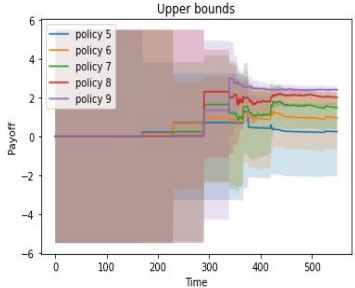

| policies | number of pulls |
|---|---|
| policy 5 | 0 |
| policy 6 | 2 |
| policy 7 | 0 |
| policy 8 | 20 |
| policy 9 | 170 |

Figure 4: Evolution of confidence bounds for the top 5 policies.

Figure 5: *Number of pulls for the top 5 policies, $\sigma = 1$.*

The correlation between policies is controlled by the values of $\sigma$. Figure 6a shows the cumulative regret of COPTIMIST for different values of $\sigma$. High values of $\sigma$ improve the convergence speed since playing a policy gives informative feedback to other policies. The evolution of the budget is plotted in figure 6b.

### E.2 Mountain Car

**Bonus clipping** The experiments showed in the main body were done with a clipped threshold $|\beta_k(\pi)| < 20$. We started by clipping at the higher value of 100 Fig. 7. According to Fig. 8, it corresponds to a lower bound of the difference between the RBH estimates and its associated bonus. While the agent eventually converges towards the optimal policy, it needs more than 5000 episodes.

As we can see in Fig. 7, the agent conservatively find good policies, but does not play them often. Considering Fig. 8, we can see the bounds after 5000 episodes are close to 100, leading to a bonus always clipped at 100. Instead of increasing the length of our experiments, we preferred restricting the pessimistic bonus to be 20. It handles the uncertainty on the RBH estimate, while giving the agent a chance to learn efficiently.

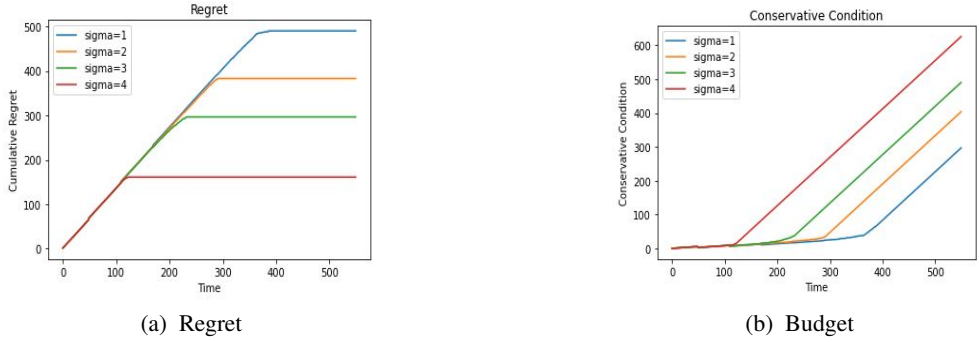

(a) Regret

(b) Budget

Figure 6: Regret and budget of COPTIMIST for different values of $\sigma$.

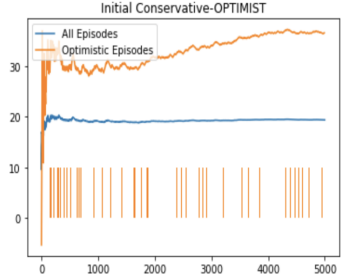

Figure 7: Mean Return of our initial C-OPTIMIST 2 on Mountain Car

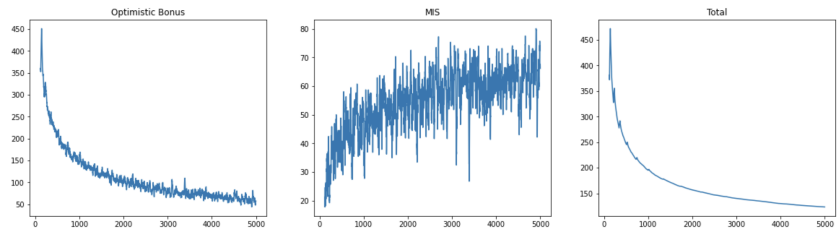

Figure 8: MIS and Optimistic bonus of OPTIMIST

**C-UCBVI** We apply UCBVI and C-UCBVI to a discretized version of Continuous MountainCar. While the comparison is unfair as C-UCBVI has access to a much wider range of policies, including close-to-optimal ones, we disclose it here for completeness.

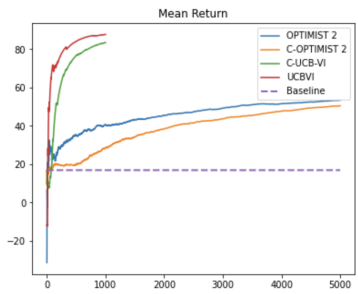

(a) Mean Return

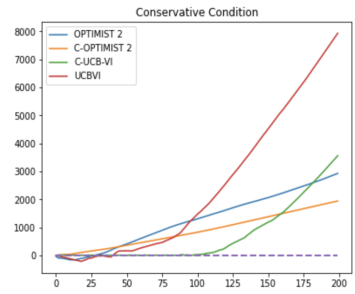

(b) Budget

Figure 9: Comparison of Conservative agents